\documentclass[journal]{IEEEtran}

\ifCLASSINFOpdf
   \usepackage[pdftex]{graphicx}
\else
   \usepackage[dvips]{graphicx}
\fi

\usepackage{amsmath}
\usepackage{amssymb}
\usepackage{mathrsfs}
\usepackage{algorithm}
\usepackage{algorithmic}
\usepackage{subfigure}
\usepackage{xcolor}
\usepackage{bm}
\usepackage{booktabs}
\usepackage{threeparttable}
\usepackage{multirow}
\usepackage{algorithm}
\usepackage{algorithmic}
\usepackage[numbers,sort&compress]{natbib}

\usepackage[colorlinks, citecolor=black]{hyperref}
\usepackage[numbers,sort&compress]{natbib}
\usepackage{array}
\usepackage{graphicx}
\usepackage{extarrows}
\graphicspath{{figures/}}

\begin{document}

\newcommand{\secref}[1]{Section~\ref{sec:#1}}
\newcommand{\figref}[1]{Figure~\ref{fig:#1}}
\newcommand{\tabref}[1]{Table~\ref{tab:#1}}
\newcommand{\eqnref}[1]{Eq.~\eqref{eq:#1}}
\newcommand{\algref}[1]{Algorithm~\ref{alg:#1}}

\newenvironment{myitemize2}[1][]{
\begin{list}{$\bullet$}
    {
     \setlength{\leftmargin}{5mm} 
     \setlength{\parsep}{0.5mm} 
     \setlength{\topsep}{0mm} 
     \setlength{\itemsep}{0mm} 
     \setlength{\labelsep}{0.5em} 
     \setlength{\itemindent}{0mm} 
     \setlength{\listparindent}{6mm} 
    }}
{\end{list}}

\title{Multimodal Imbalance-Aware Gradient Modulation for Weakly-supervised Audio-Visual Video Parsing}


\author{Jie~Fu,
	Junyu~Gao,
	and~Changsheng~Xu,~\IEEEmembership{Fellow,~IEEE}
	\IEEEcompsocitemizethanks{Jie Fu is with Zhengzhou University, ZhengZhou 450001, China, and also with the State Key Laboratory of Multimodal Artificial Intelligence Systems (MAIS), Institute of Automation, Chinese Academy of Sciences, Beijing 100190, China. (email: fujie@gs.zzu.edu.cn).
	
	Junyu Gao and Changsheng Xu are with the State Key Laboratory of Multimodal Artificial Intelligence Systems (MAIS), Institute of Automation, Chinese Academy of Sciences, Beijing 100190, P. R. China, and with School of Artificial Intelligence, University of Chinese Academy of Sciences, Beijing, China. Changsheng Xu is also with Peng Cheng Laboratory, ShenZhen 518055, China. (e-mail: junyu.gao@nlpr.ia.ac.cn; csxu@nlpr.ia.ac.cn).}
	
}

\markboth{Journal of \LaTeX\ Class Files,~Vol.~14, No.~8, August~2015}%
{Shell \MakeLowercase{\textit{et al.}}: Bare Demo of IEEEtran.cls for IEEE Journals}

\maketitle

\begin{abstract}
Weakly-supervised audio-visual video parsing (WS-AVVP) aims to localize the temporal extents of 
audio, visual and audio-visual event instances as well as identify the corresponding event categories 
with only video-level category labels for training. Most previous methods pay much attention to 
refining the supervision for each modality or extracting fruitful cross-modality information for more reliable 
feature learning. None of them have noticed the imbalanced feature learning between different modalities in the task. 
In this paper, to balance the feature learning processes of different modalities, a dynamic gradient 
modulation (DGM) mechanism is explored, where a novel and effective metric function is designed to 
measure the imbalanced feature learning between audio and visual modalities.
Furthermore, principle analysis indicates that the multimodal confusing calculation 
will hamper the precise measurement of multimodal imbalanced feature learning, which further
weakens the effectiveness of our DGM mechanism. To cope with this issue, a modality-separated 
decision unit (MSDU) is designed for more precise measurement of imbalanced feature learning between 
audio and visual modalities.
Comprehensive experiments are conducted on public benchmarks and the corresponding experimental
results demonstrate the effectiveness of our proposed method. 
\end{abstract}


\IEEEpeerreviewmaketitle

\section{Introduction}
\label{sec:intro}


Recently, many different audio-visual video understanding tasks such as audio-visual action recognition~\cite{planamente2022domain, 
gao2020listen, kazakos2019epic},
audio-visual separation~\cite{tzinis2022audioscopev2, gan2020music, 
tzinis2020into, tian2021cyclic, gao2021visualvoice} and audio-visual event localization~\cite{tian2018audio, wu2019dual, xuan2020atten, duan2021audio, 
xue2021audio, xu2020cross} have been proposed and achieve impressive
progresses. The above audio-visual parsing models are all learned based on the assumption that
audio and visual modalities are temporally aligned and the corresponding fine-grained frame-level
annotations of different modalities are also provided for training. 
\begin{figure}[thbp]
  \centering
  \includegraphics[width=1\linewidth]{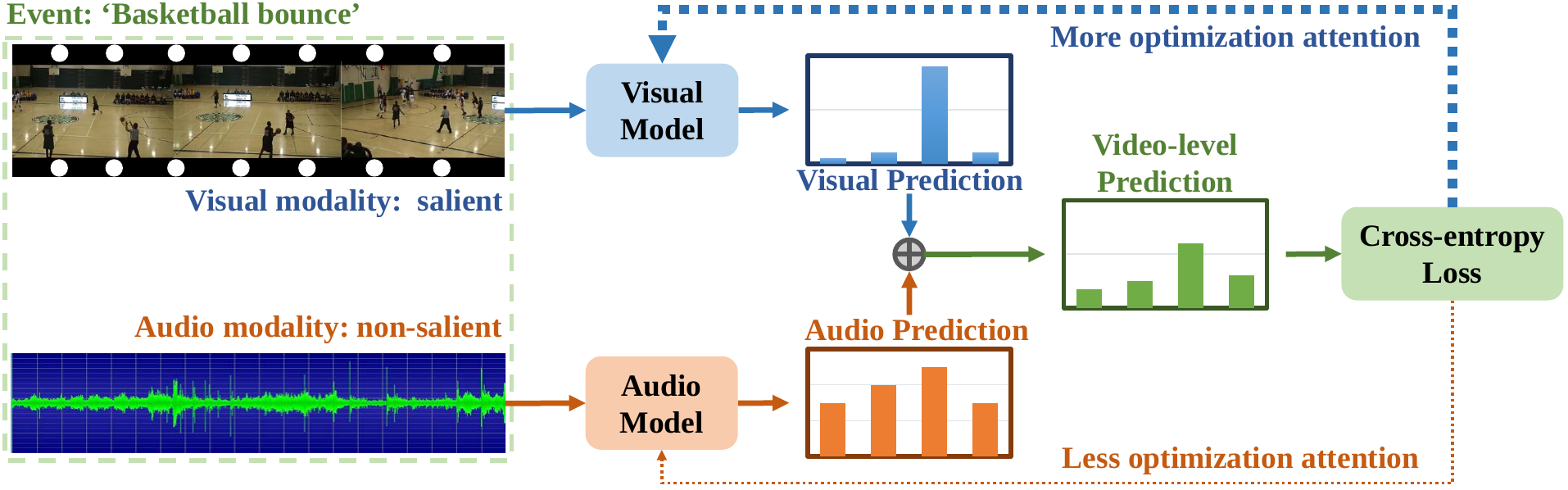}
  \caption{Simple illustration of our motivation. During model training, the modality containing obvious semantic information will dominate the training progress 
  and achieve more optimization attention, leading to suboptimal issue in the other modality.
  Consequently, we attempt to balance the multimodal feature learning by mudulating gradients.}
  \label{fig:motivation}
\end{figure}
However, in real-world scenes, 
the synchronization between audio and 
visual modalities is not always satisfied and annotating frame-level labels for massive videos is 
time-consuming and unfeasible. 
To mitigate the above issues, weakly-supervised audio-visual
video parsing task~\cite{tian2020unified} is proposed. In the task, only video-level event 
category labels are annotated for training the model, which attempts to 
identify the starting and ending timestamps of each event instance and predict the corresponding event 
categories in terms of modalities (\emph{i.e.}, audio, visual, or both) during the inference stage.

Inspired by the prior knowledge that multiple modalities can provide more effective information from 
different aspects than only single one, most existing WS-AVVP methods~\cite{tian2020unified, 
lin2021exploring, cheng2022joint} are designed in the same pipeline 
to aggregate multimodal information for more precise audio-visual video parsing:
Firstly, the HAN cross-attention~\cite{tian2020unified} is utilized to enhance the audio and visual feature
representation. Afterwards, the enhanced feature of different modalities are input 
into the same attention module and classification head to generate the modality-specific attention weights and classification scores, 
which are aggregated to produce the 
final video-level classification predictions. Throughout the training process, a uniform learning objective and joint 
training strategy are utilized to optimize the sub-networks of different modalities.

By going in depth into the pipeline mentioned above, an intuitive deficiency can be 
concluded that the audio and visual modalities are optimized equally and the 
natural discrepancy between them is overlooked.
Concretely, as shown in~\figref{motivation}, it is difficult to judge whether `Basketball bounce' is happening by 
just listening to the audio stream, but it is obvious enough in visual data. 
Consequently, during the training process, the modality conveying more salient semantic information will
dominate the training process and obtain more optimization attention, and the modality containing 
relatively confusing information will not be fully optimized~\cite{parida2020coordinated}.
Ultimately, variant modalities often trend to convergence 
at different rates, which will further result in uncoordinated convergence issue~\cite{ismail2020improving, 
sun2021learning, wang2020makes}.

To cope with the above issue, a \textbf{Dynamic Gradient Modulation (DGM)} mechanism is explored to 
balance the feature learning processes of audio and visual modalities. Concretely, a novel 
metric function is first designed to measure the imbalanced feature learning degree between 
different modalities, which followed by utilizing the imbalance degree to modulate the backward 
gradients of the sub-networks for different modalities, so as to drive the AVVP model to pay 
more optimization attention to the suboptimal modality. So far, a similar imbalance metric function~\cite{peng2022balanced} 
based on the predicted scores of the correct categories in different modalities has been proposed. 
However, it neglects the global prediction distribution information of all categories, 
which is also beneficial for precise imbalance assessment. Consequently, in our DGM, a more 
thoughtful imbalance 
metric function considering both above two information is designed.


Furthermore, in traditional WS-AVVP pipeline, the cross-attention operation is always utilized to
exchange audio and visual information, which will lead to the confusing multimodal
information. In this case, it is hard to measure the imbalanced feature learning in different modalities 
purely, which will further damage the effectiveness of our proposed DGM.
In order to address the above issue, according to the principle analysis, 
we design a~\textbf{modality-separated decision unit (MSDU)} structure,
which is embeded between the modality-specific feature encoders and cross-attention block of the traditional pipeline
for more precise measurement of imbalanced feature learning between different modalities.
In MSDU, the 
calculation of different modalities is separated completely, which is proved to be beneficial for 
enlarging the performance gains brought by our DGM mechanism.


Ultimately, to evaluate the performance  of our proposed method, we conduct 
extensive comparison and ablation studies on several widely utilized audio-visual benchmarks. 
Experimental results show that our proposed method 
achieves the state-of-the-art performance. To summarize, our main contributions are three-fold:
\begin{myitemize2}
      \item 
      We analyze the imbalanced feature learning issue in WS-AVVP task. To this end, a dynamic 
      gradient modulation mechanism is proposed to modulate the gradients of variant sub-networks
      for different modalities according to their contributions to the learning objective, so as to 
      make the multimodal framework pay more attention to the suboptimal modality.

      \item
      We observe that the confusing calculation of different modalities will disturb the precise measurement
      of multimodal imbalanced feature learning, which will further
      damage the effectiveness of our proposed DGM mechanism, thus we design a 
      Modality-Separated Decision Unit, which can cooperate with the 
      proposed DGM for a significant improvement.
  %
  
      \item
      Comprehensive experimental results show that our proposed model outperforms all current 
      state-of-the-arts, which verifies the effectiveness of our proposed DGM mechanism and MSDU structure.
\end{myitemize2}


\section{Related Work}
\label{sec:related}
In this section, we review the most related work to our method including audio-visual learning and understanding as well as weakly-supervised audio-visual video parsing.

\subsection{Audio-visual learning and understanding}
As the two most common and fundamental sensory information, visual and auditory data have attracted a large 
amount of research attention and derived many different audio-visual understanding tasks, such as 
audio-visual feature representation learning~\cite{aytar2016soundnet, hu2019deep, korbar2018cooperative, arandjelovic2017look, 
cheng2020look, ma2020active, morgado2020learning}, audio-visual action recognition~\cite{planamente2022domain, 
gao2020listen, kazakos2019epic}, sound source localization~\cite{rachavarapu2021localize, xuan2022proposal, 
owens2018audio, senocak2018learning}, audio-visual video captioning~\cite{hori2017attention, rahman2019watch, 
tian2018attempt}, audio-visual event localization~\cite{tian2018audio, wu2019dual, xuan2020atten, duan2021audio, 
xue2021audio, xu2020cross} and audio-visual sound separation~\cite{tzinis2022audioscopev2, gan2020music, 
tzinis2020into, tian2021cyclic, gao2021visualvoice}. Most audio-visual learning methods are designed based 
on the assumption that audio and visual modalities are synchronized and temporal correlated. Concretely, 
\cite{cheng2020look, korbar2018cooperative, ma2020active, morgado2020learning, owens2018audio} attempt to 
learn audio-visual feature representation jointly by utilizing the temporal alignment information between
audio and visual modalities as the self-supervised guidance. In~\cite{arandjelovic2017look},
a novel pretext task is proposed to extract correspondent feature for correlated video frame and audio, 
where the audio-visual feature pairs belonging to the same temporal snippet are gathered and the features 
from unpaired video snippets are separated. In~\cite{hu2019deep}, the unsupervised multimodal 
clustering information is utilized as the supervision for cross-modality feature correlation learning. 
To enhance the object detection capability, the correlation information between audio and object motion is 
taken into consideration~\cite{gan2020music, gan2019self, zhao2019sound, zhao2018sound}.
The methods mentioned above have achieved some progress by utilizing the synchronization
between different modalities, which is not always satisfied in realistic scenes. 

\subsection{Weakly-supervised audio-visual video parsing}

WS-AVVP aims to parse an arbitrary untrimmed video into a group of event
instances associated with semantic categories, temporal boundaries and modalities (\emph{i.e.}, audio, visual 
and audio-visual), where only the video-level labels are provided as the supervision information
for training. Tian~\emph{et al.}~\cite{tian2020unified} first propose the WS-AVVP setting and design a hybrid attention
network in a multimodal multiple instance learning (MMIL) pipeline, where the intra- and cross-modality
attention strategies are utilized to capture cross-modality contextual information. 
Afterwards, Lin~\emph{et al.}~\cite{lin2021exploring} further explore cross-video and 
cross-modality complementary information to facilitate WS-AVVP, where both the common and diverse event semantics 
across videos are exploited to identify audio, visual and audio-visual events. Thereafter, to refine the event
labels individually for each modality, Wu~\emph{et al.}~\cite{wu2021exploring} propose a novel method by swapping
audio and visual modalities with other unrelated videos. JoMoLD~\cite{cheng2022joint} utilizes the cross-modality 
loss pattern to help remove the noisy event labels for each modality. Although the methods mentioned above have enhanced 
the cross-modality feature learning and mitigated the issue of modality-specific noisy labels, none of them have 
noticed the imbalanced multimodal feature learning. 

\subsection{Imbalanced audio-visual learning}

Compared with the uni-modality learning, multimodal learning can integrate more information from different
aspects.
However, there always exists a discrepancy between 
different modalities, which makes the extraction and fusion of information from different modalities 
become more challenging. Concretely, the information volume and complexity in different modalities are often 
variant, thus the training difficulty of the networks for variant modalities is also different, which will 
lead to uncoordinated optimization issue~\cite{du2021improving, ismail2020improving, sun2021learning, wang2020makes}: 
the dominated modality~\cite{parida2020coordinated} conveying salient information
will achieve more optimization attention and better performance during the whole training process, which will
suppress the optimization progress of the other one. 
To cope with the above issue, Wang~\emph{et al.}~\cite{wang2020makes} propose a metric to measure the overfitting-to-generalization ratio (OGR) and design a novel training scheme to minimize the OGR via an optimal blend of 
multiple supervised signals. To enhance the feature embedding capability of the suboptimal modality, 
Du~\emph{et al.} propose to distill reliable knowledge from the well-optimized model to help strengthen
the optimization of the other one. Similar to our work, OGM~\cite{peng2022balanced} proposes to 
measure the optimization discrepancy between different modalities by calculating the ratio of predicted scores 
for the correct categories in different modalities, which is then utilized to modulate the 
gradients of different modality-specific models. However, OGM only takes the prediction scores for the correct
event categories into consideration, which is not sufficient for reliable imbalance assessment.
Consequently, we further consider the global prediction distribution on all categories.
Meanwhile, to mitigate the negative effect of multimodal confusing calculation
on our proposed DGM mechanism, we design the MSDU, which can cooperate with DGM for more significant improvement.
\section{Our approach}
\label{sec:method}
\begin{figure*}
    \centering
    \includegraphics[width=0.9\linewidth]{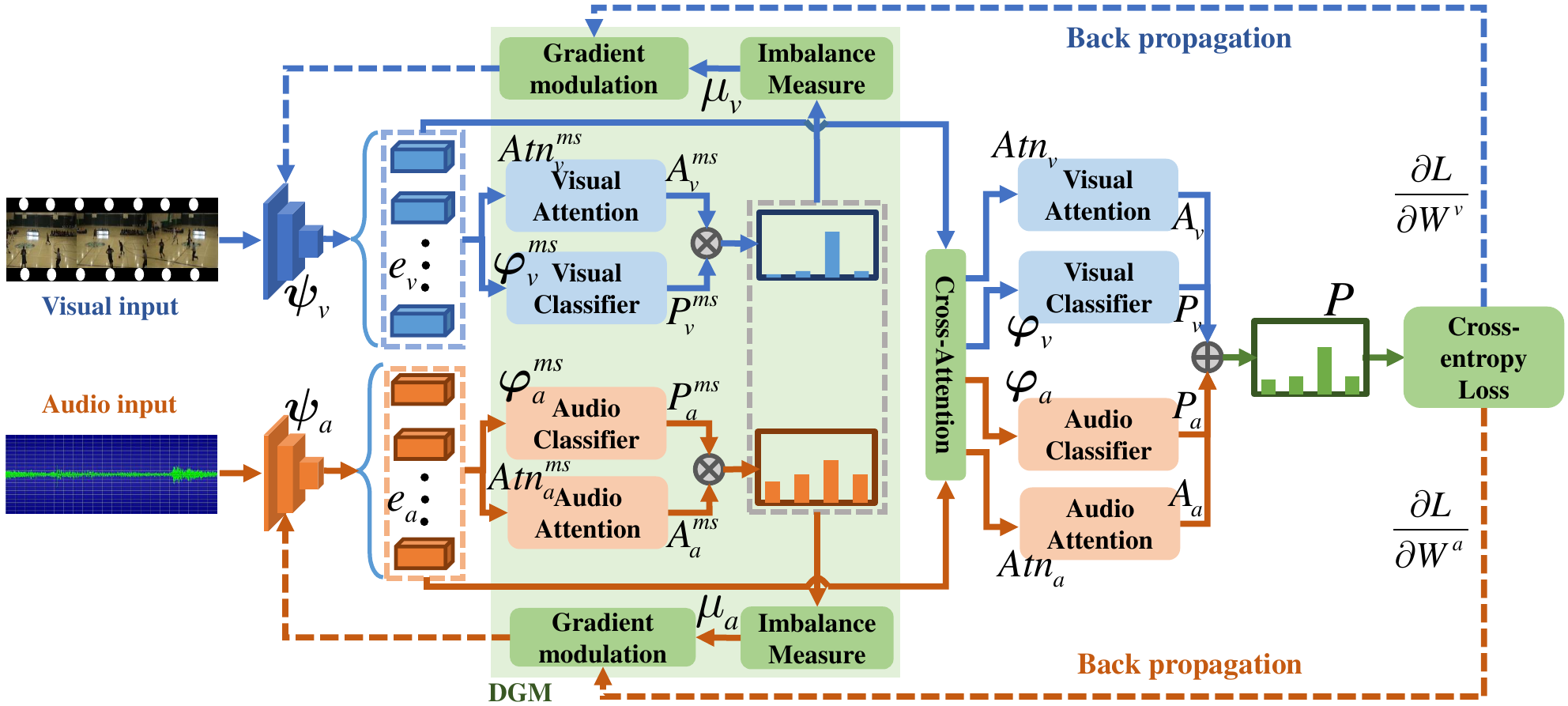}
    \caption{Overview of our proposed framework. DGM  
    is utilized to modulate the gradients of modality-specific feature encoders  
    according to their contributions to the learning objective, so as to balance the 
    feature learning of different modalities. The MSDU structure is proposed to separate 
    the calculation of different modalities, which can cooperate with DGM for 
    more significant improvement.}
    \label{fig:framework}
\end{figure*}

\subsection{Problem Formulation}
In this paper, we follow the standard protocol of WS-AVVP ~\cite{tian2020unified}: 
Formally, given 
a multimodal training video sequence $\left\{ a^{t}, v^{t} \right\}_{t=1}^{T}$ 
with $T$ snippets, only the coarse video-level label $Y\in \left\{ 0, 1\right\}^{C}$ is 
available during the training process, where $a^{t}$ and $v^{t}$ indicate the $t$-th audio and visual 
snippets and $C$ denotes the number of event categories.
In the testing phase, the trained model should predict the snippet-level semantic category label 
$\mathbf{y}_{t}=\left\{ y_{t}^{a}, y_{t}^{v}, y_{t}^{av} \right\}$ for audio, visual and audio-visual 
modalities respectively, where $y_{t}^{av}=y_{t}^{a}\times y_{t}^{v}$. In other words, the audio-visual events
happen only when the corresponding audio and visual events belonging to the same semantic category occurr at the same time. 
Notably, the video-level event category label (multi-hot vector) can be obtained by aggregating 
the snippet-level event category annotations along the temporal dimension in each video.
Due to the lack of fine-grained snippet-level supervision information during the training 
process, current WS-AVVP models are all designed in the multimodal multiple instance learning (MMIL) pipeline:
in the training process, the snippet-level predictions from different modalities in the same video
are aggregated to form the video-level prediction, which is supervised by the annotated video-level label.

\subsection{Optimization Analysis and Our DGM}
The pipeline of our proposed framework is illustrated in~\figref{framework}:  
Compared with our proposed framework, the traditional WS-AVVP pipeline (\emph{i.e.}, our proposed framework without 
the component in the light green background box) contains three main components including audio and visual 
feature encoders $\psi_{a/v}\left( \cdot \right)$ for snippet-level video feature learning, cross-attention 
mechanism $cro\_att(\cdot)$ for exchanging multimodal information, as well as multimodal attention 
$Atn_{a/v}\left(\cdot\right)$ and classification $\varphi_{a/v}\left(\cdot\right)$ modules for snippet-level 
video action prediction.

During the training process, an arbitrary training video containing audio $a$ and visual $v$ modalities is input into 
$\psi_{a/v}\left( \cdot \right)$ to produce the modality-specific snippet-level video 
feature, which is then fed into the cross-attention function $cro\_att(\cdot)$ to extract multimodal cross-attented 
feature representation. Afterwards, 
$Atn_{a/v}\left(\cdot\right)$ and $\varphi_{a/v}\left(\cdot\right)$ are utilized to generate the 
modality-specific attention weights and classification predictions. Ultimately, the attention weights and classification 
predictions in different modalities are combined together to produce 
the video-level action classification prediction, which is supervised by the video-level labels provided by the annotators.
To optimize the model, the optimization objective of the traditional pipeline can be 
formulated as follows:
\begin{equation}
    \begin{aligned}
        L = -\frac{1}{N C} \sum_{n=1}^{N}\sum_{c=1} ^{C} [Y_{n}[c]\log &P_{n}[c] +  \\
        (1 - &Y_{n}[c]) \log (1 - P_{n}[c])]
    \end{aligned}
\end{equation}
where $N$ denotes the number of training samples in a mini-batch and $Y\left[ c \right]$ indicates the $c$-th element
of $Y$. $P_{n}$ is the classification prediction for the $n$-th video, which is obtained via sigmoid function.
Concretely, video-level classification prediction $P_{n}$ can be formally calculated as follows:
\begin{equation}
    \begin{aligned}
        P_{n} &= \varphi\left( f^{n}_{av} \right) = \alpha^{w} [f^{n}_{a}, f^{n}_{v}] + [\alpha^{b}_{a} , \alpha^{b}_{v}]    \\
          &= \alpha^{w}_{a}\cdot f^{n}_{a} + \alpha^{b}_{a} + \alpha^{w}_{v}\cdot f^{n}_{v} + \alpha^{b}_{v}   \\
          &= \alpha^{w}_{a}\cdot \pi_{a}(a_{n}; \theta_{a}) + \alpha^{b}_{a} + \alpha^{w}_{v}\cdot \pi_{v}(v_{n}; \theta_{v}) + \alpha^{b}_{v}
    \end{aligned}
\end{equation}
where $\alpha^{w}$ and $\alpha^{b}$ denote the weight and bias of modality-specific classifiers, 
$\pi(\cdot)$ is the formalization of feature embedding function including the modality-specific feature 
encoder $\psi(\cdot)$ and cross-attention block $cro\_att(\cdot)$, 
and $\theta$ indicates the hyper-parameter of $\pi(\cdot)$. In the following content, we omit the bias 
matrix $\alpha^{b}$ for more brevity.
During the optimization process of our model, the gradient descent strategy is utilized to update the parameters
of our model. Formally, the optimization processes of different parameters in our model can be formulated 
as follows (The superscript and subscript $a / v$ are omitted for brevity):
\begin{equation}
    \begin{aligned}
        \alpha^{w} &\gets \alpha^{w} + \lambda \frac{\partial{L}}{\partial{\varphi\left( f_{av} \right)}} \frac{\partial{\varphi\left( f_{av} \right)}}{\partial{\alpha^{w}}}    \\
                         &= \alpha^{w} + \lambda \frac{1}{N} \sum_{n=1}^{N} \frac{\partial{L}}{\partial{\varphi\left( f^{n}_{av} \right)}}\cdot f^{n}
    \end{aligned}
\end{equation}
\begin{equation}
    \begin{aligned}
        \theta &\gets \theta + \lambda \frac{\partial{L}}{\partial{\varphi\left( f_{av} \right)}} \frac{\partial{\varphi\left( f_{av} \right)}}{\partial{\theta}} \\
                     &= \theta + \lambda \frac{1}{N} \sum_{n=1}^{N} \frac{\partial{L}}{\partial{\varphi\left( f^{n}_{av} \right)}} \frac{\partial{\varphi\left( f^{n}_{av} \right)}}{\partial{\theta}}
    \end{aligned}
\end{equation}
where $\lambda$ is the learning rate of the optimizer. Obviously, the common item in the above two equations is (Refer to \textbf{Supplementary Material} for detailed deduction of~\eqnref{grad}): 
\begin{equation}
    \begin{aligned}
        \frac{\partial{L}}{\partial{\varphi\left( f_{av}^{n} \right)}}[c] = \frac{1}{1 + e^{- \varphi\left( f_{av}^{n} \right)}}[c] - Y_{n}[c] \\
        = \frac{1}{1 + e^{- [\alpha^{w}_{a}\cdot \pi_{a}(a_{n}; \theta_{a}) + \alpha^{w}_{v}\cdot \pi_{v}(v_{n}; \theta_{v})]}}[c] - Y_{n}[c]
    \end{aligned}
    \label{eq:grad}
\end{equation}
From the above formulations, we can intuitively draw 
the conclusion that if the semantic information contained in the audio modality is much more obvious than the 
visual one, the audio will achieve higher predicted confidence scores. Furthermore, the back-propagation gradient
(\eqnref{grad}) is more contributed by $\alpha^{w}_{a}\cdot \pi_{a}(a_{n}; \theta_{a})$, which
will make the audio achieve more optimization attention. 
Consequently, the visual modality will have a relatively lower prediction confidence and 
limited optimization efforts will be paid to it during the training process. Ultimately, although
the training process of the whole multimodal model has converged, the modality containing relatively weak semantic information
could not be fully optimized.

To cope with this issue, we propose a simple but effective dynamic gradient modulation (DGM) strategy (\emph{i.e.}, the component 
shown in the light green background box of~\figref{framework})
to balance the feature learning of audio and visual modalities. 
Specifically, inspired by the factor that fully optimized model will predict higher classification scores 
for those correct categories and the discrepancy between the predicted scores for correct and wrong categories will be
large, we assess the relative optimization progress between visual and audio modalities as follows:
\begin{equation}
    \begin{aligned}
        \omega_{v-a} = \frac{\sum_{n} \sum_{c} s_{n}^{v}[c] \cdot Y_{n}[c] + \sum_{n} \triangle \bar{s}_{n}^{v}}
                              {\sum_{n} \sum_{c} s_{n}^{a}[c] \cdot Y_{n}[c] + \sum_{n} \triangle \bar{s}_{n}^{a}}
    \end{aligned}
    \label{eq:mu}
\end{equation}
where $s_{n}^{v}$ and $s_{n}^{a}$ are visual and audio classification confidence scores of the $n$-th sample, 
and $\triangle \bar{s}_{n}$ denotes the corresponding discrepancy between the average prediction scores of the correct
and wrong event categories. Similarly,
$\omega_{a-v}$ is the reciprocal of $\omega_{v-a}$. If the $\omega_{v-a}$ is larger than $1$, the optimizer will pay more attention to the visual modality,
which will result in suboptimal audio feature learning. As a result, the balance coefficients of different modalities 
can be designed as follows:
\begin{equation}
    \begin{aligned}
        \mu^{v} = \left\{\begin{matrix}
            1 - \tanh  \left(\gamma \cdot \omega_{v-a}\right), & if \quad \omega_{v-a} > 1 \\
            1, & if \quad \omega_{v-a} \le 1
          \end{matrix}\right.
    \end{aligned}
\end{equation}
where $\tanh \left(\cdot\right)$ denotes the activation function and $\gamma$ is a hyper-parameter managing the modulation degree. 
$\mu^{a}$ can be obtained in a similar way, but we omit the corresponding description for clarity.
Afterwards, we utilize the balance coefficients $\mu$ to modify the gradients of different sub-networks
during the back-propagation process:
\begin{equation}
    \vspace{-2mm}
    \begin{aligned}
        W \gets W + \lambda \cdot \mu \cdot \frac{\partial{L}}{\partial{W}}
    \end{aligned}
\end{equation}
where $W$ indicates the parameter to be optimized (\emph{i.e.,} $\theta$ and $\alpha^{w}$ in our model).

Moreover, according to~\cite{mandt2017stochastic, jastrzkebski2017three, peng2022balanced}, the back-propagation gradients in each batch follow a Gaussian distribution and an appropriately large gradient covariance will lead to better generalization 
ability. However, the gradient covariance modified by our DGM mechanism becomes $\mu^{2} \cdot \sigma^{2}(\frac{\partial{L}}{\partial{W}})$, 
which is smaller than the original one $ \sigma^{2} (\frac{\partial{L}}{\partial{W}})$ because $\mu \in (0, 1]$. 
To make up for this deficiency, we add an extra term to the modified gradient covariance as follows:
\begin{equation}
    \begin{aligned}
        W &\gets W + \lambda \cdot \mathrm {E}(\frac{\partial{L}}{\partial{W}}) + \lambda \varepsilon,     \\
        where \quad \varepsilon & \sim \mathcal{N}(0, (\mu^{2} + 1) \cdot \sigma^{2}(\frac{\partial{L}}{\partial{W}}))
    \end{aligned}
    \label{eq:gc}
\end{equation}
where $E(\cdot)$ and $\sigma^{2}(\cdot)$ are the expectation and covariance.


\begin{figure}[thbp]
    \subfigure[Traditional pipeline.]{
        \includegraphics[width=0.98\linewidth]{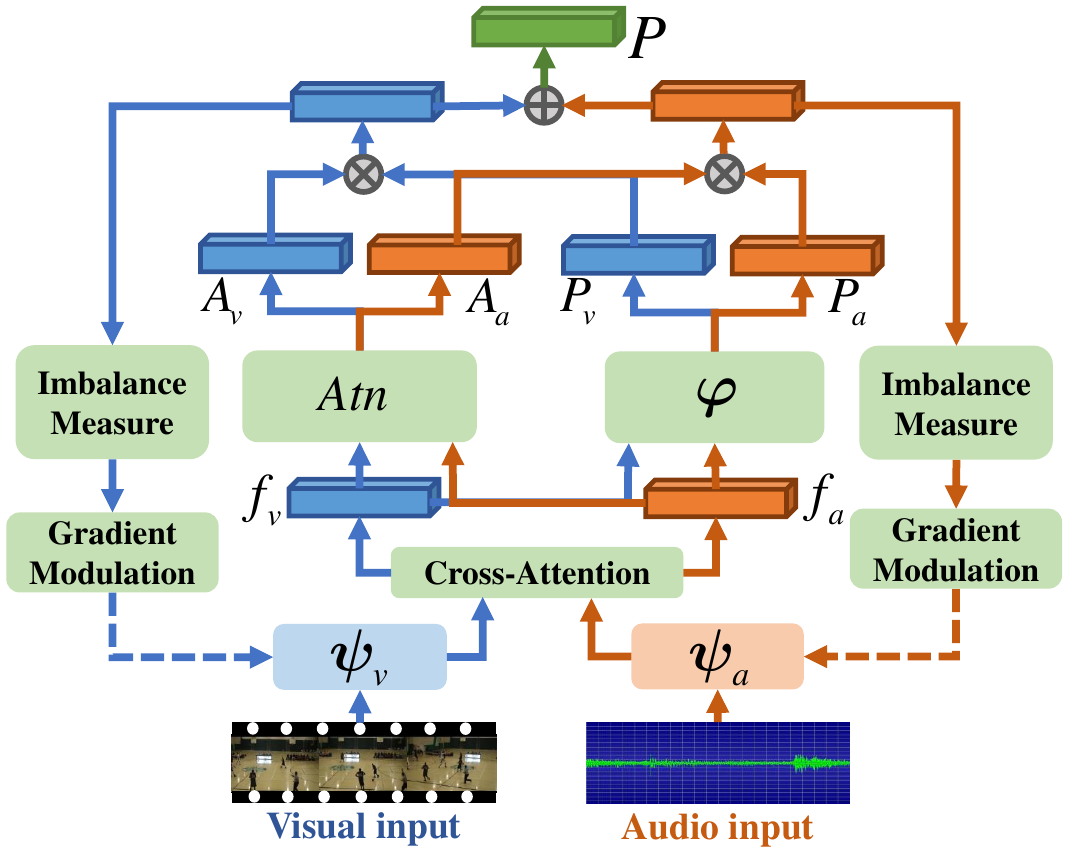}
        \label{fig:mot1}
    }
    \subfigure[Our proposed framework pipeline.]{
        \includegraphics[width=0.98\linewidth]{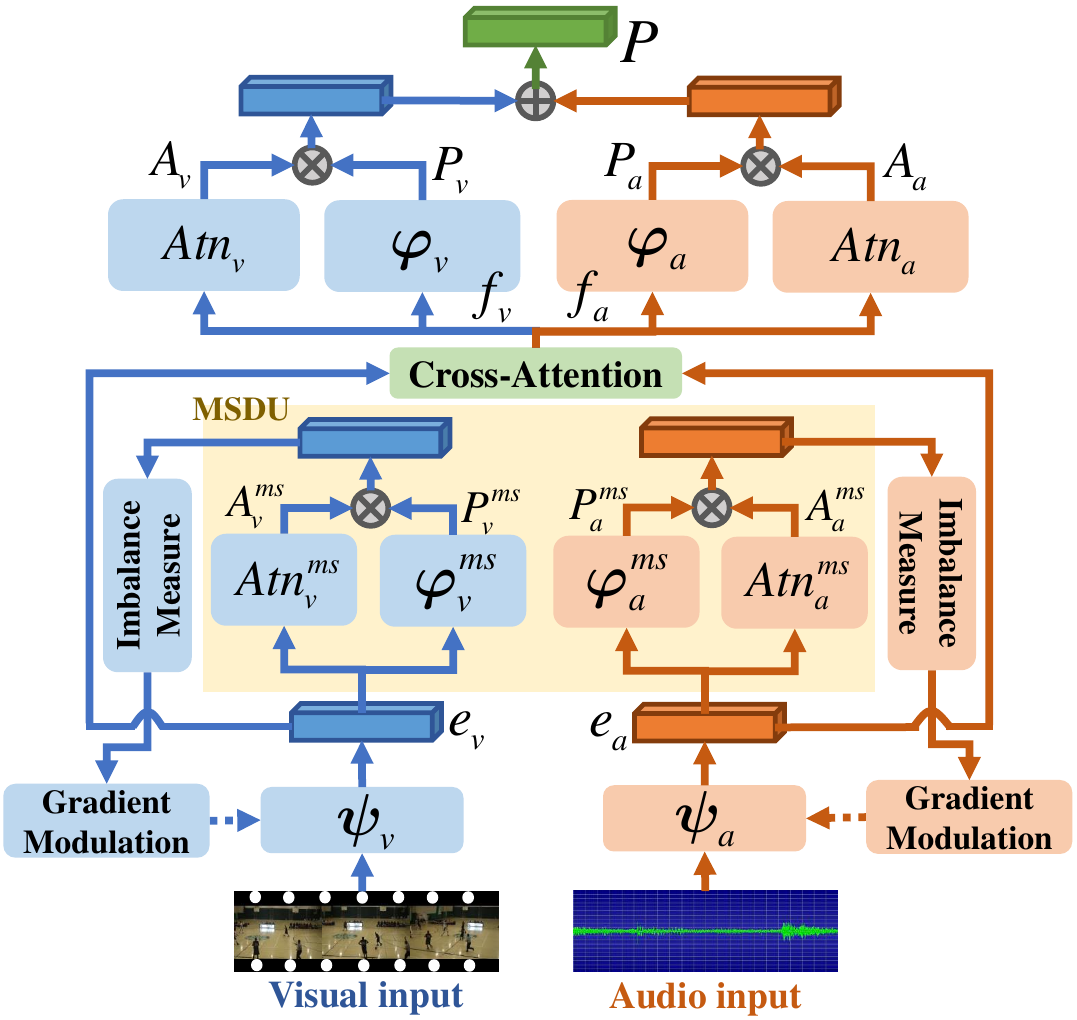}
        \label{fig:mot2}
    }
    \caption{Traditional pipeline \emph{vs.} our proposed framework.}
    \label{fig:trad_safe}
\end{figure}

\subsection{Modality-Separated Imbalance Measurement}
Obviously, in our proposed DGM mechanism, the core idea is to utilize the discrepancy between the predictions of different modalities
to assess the imbalanced feature learning between audio and visual modalities, which is further applied to modulate the gradient 
of each modality-specific feature encoder during the optimization process. 
However, almost all existing WS-AVVP models~\cite{tian2020unified, lin2021exploring, wu2021exploring, 
cheng2022joint} are designed in the following pipeline as shown in~\figref{trad_safe}(a):
audio $a$ and visual $v$ data are first input into the feature encoders $\psi_{a/v}\left( \cdot \right)$ to produce 
the modality-specific feature. 
Thereafter, the cross-modality attention mechanism $cro\_att(\cdot)$ is utilized to 
exchange the related information in different modalities and produced the cross-attented features 
$f_{a}$ and $f_{v}$.
Afterwards, $f_{a}$ and $f_{v}$ are separately fed into the classifier $\varphi(\cdot)$ and attention module 
$Atn(\cdot)$ shared by audio and visual modalities to produce the snippet-level classification probabilities 
(\emph{i.e.}, $P_{a}$ and $P_{v}$) 
and temporal attention weights (\emph{i.e.}, $A_{a}$ and $A_{v}$) for different modalities. Ultimately, 
$P_{a}$, $P_{v}$, $A_{a}$ and $A_{v}$ are aggregated together to generate the video-level event classification
prediction $P$, which is supervised by the video-level event category label provided by the annotators.
Formally, the above pipeline can be written as follows:
\begin{equation}
    \vspace{-2mm}
    \begin{aligned}
        f_{a}, \ f_{v} &= cro\_att( \psi_{a}(a), \ \psi_{v}(v)) \\
        P_{a}, \ P_{v} &= \varphi(f_{a}), \ \varphi(f_{v}),  \\
        A_{a}, \ A_{v} &= Atn(f_{a}), \ Atn(f_{v})   \\
        P = A&gg_{T}(A_{a} * P_{a} + A_{v} * P_{v})
    \end{aligned}
    \label{eq:tradition}
\end{equation}
where $`*'$ and $`+'$ denote the broadcast multiplication and element-wise addition, and $Agg_{T}(\cdot)$ is the aggregation operation along the temporal dimension.
In the traditional case, $P_{a/v}$ and $A_{a/v}$ are utilized to calculate $s^{a/v}$ and $\triangle \bar{s}^{a/v}$ 
for the assessment of imbalanced feature learning between different modalities.

By going in depth into the traditional pipeline mentioned above, we can find that due to the cross-attention operation 
between audio and visual modalities, both $f_{a}$ and $f_{v}$ encode the information from different modalities.
Consequently, the information of different modalities are confusing, which 
makes it hard to purely measure the imbalanced feature learning between different modalities by directly
utilizing the attention $A_{a/v}$ and classification predictions $P_{a/v}$ produced based on the confusing multimodal 
feature $f_{a/v}$. 

According to the above analysis, we modify the traditional pipeline and 
propose a modality-separated decision unit (MSDU) for more pure assessment of imbalanced feature learning between 
different modalities. 
Concretely, MSDU is embeded between the modality-specific feature encoders $\psi_{a/v}\left( \cdot \right)$ and 
cross-attention block $cro\_att(\cdot)$. Our proposed pipeline is shown in~\figref{trad_safe}(b). 
Obviously, in our pipeline, apart from our proposed MSDU (\emph{i.e.}, the component shown in yellow background box), 
the rest is similar to the traditional WS-AVVP pipeline: audio $a$ and visual $v$ data is first input into the modality-specific 
feature encoders to produce the modality-specific video feature $e_{a/v}$, which are then fed into the cross-attention module for more 
robust feature extraction. Afterwards, the cross-attented multimodal feature $f_{a}$ and $f_{v}$ are further taken 
as the inputs of modality-specific attention blocks $Atn_{a/v}(\cdot)$ and classification modules $\varphi_{a/v}(\cdot)$, 
which produce snippet-level action attention weights $A_{a/v}$ and classification scores $P_{a/v}$.
To achieve more pure measurement of imbalanced multimodal feature learning, in our proposed MSDU, another group of 
modality-specific attention blocks $Atn_{a/v}^{ms}(\cdot)$ and classification modules $\varphi_{a/v}^{ms}(\cdot)$ are 
designed to predict the snippet-level action attentions $A_{a/v}^{ms}$ and classification scores $P_{a/v}^{ms}$ based on 
pure audio and visual modality features $e_{a/v}$ respectively. 
Formally, the pipeline of our method can be written as follows:
\begin{equation}
    \begin{aligned}
        e_{a}, \ e_{v} &= \psi_{a}(a), \ \psi_{v}(v) \\
        P_{a}^{ms}, \ P_{v}^{ms} &= \varphi_{a}^{ms}(e_{a}), \ \varphi_{v}^{ms}(e_{v}),   \\
        A_{a}^{ms}, \ A_{v}^{ms} &= Atn_{a}^{ms}(e_{a}), \ Atn_{v}^{ms}(e_{v})   \\
        f_{a}, \ f_{v} &= cro\_att( e_{a}, \ e_{v}) \\
        P_{a}, \ P_{v} &= \varphi_{a}(f_{a}), \ \varphi_{v}(f_{v}),   \\
        A_{a}, \ A_{v} &= Atn_{a}(f_{a}), \ Atn_{v}(f_{v})   \\
        P = A&gg_{T}(A_{a} * P_{a} + A_{v} * P_{v})
    \end{aligned}
    \label{eq:misa}
\end{equation}
In our pipeline, $P_{a/v}^{ms}$ and $A_{a/v}^{ms}$ are utilized to calculate $s^{a/v}$ and $\triangle \bar{s}^{a/v}$ in~\eqnref{mu}.
$P_{a/v}$ and $A_{a/v}$ are utilized as the final predcitions of our proposed model. In this manner, $P_{a/v}^{ms}$ and $A_{a/v}^{ms}$
are both based on single-modality data, which will lead to purer measurement of multimodal feature learning.
\section{Experiments}
\label{sec:exp}
\begin{table*}
    \centering
    \caption{Comparison with all the recent state-of-the-art methods on LLP. Our proposed
    model achieves the best performance on all audio-visual video parsing sub-tasks under both the segment-level 
    and event-level metrics. `Co-teaching' and `JoCoR' indicate the variants of~\cite{yu2019does, wei2020combating}, which are 
    reproduced by the authors of~\cite{cheng2022joint}. `CL' indicates the contrastive feature learning mechanism proposed in 
     MA~\cite{wu2021exploring}.}
    \resizebox{142mm}{!}{%
    \begin{tabular}{ll|ccccc|ccccc}
        \toprule[1.2pt]
        \multicolumn{2}{l|}{\multirow{2}{*}{Methods}}         & \multicolumn{5}{c|}{Segment-level} & \multicolumn{5}{c}{Event-level}   \\ \cline{3-12} 
        \multicolumn{2}{l|}{}                                 & A     & V    & AV   & Type & Event & A    & V    & AV   & Type & Event \\ \hline
        \multicolumn{2}{l|}{TALNet~\cite{wang2019comparison}}                           & 50.0  & -    & -    & -    & -     & 41.7 & -    & -    & -    & -     \\
        \multicolumn{2}{l|}{STPN~\cite{nguyen2019self}}                             & -     & 46.5 & -    & -    & -     & -    & 41.5 & -    & -    & -     \\
        \multicolumn{2}{l|}{CMCS~\cite{liu2019completeness}}                             & -     & 48.1 & -    & -    & -     & -    & 45.1 & -    & -    & -     \\ \hline
        \multicolumn{2}{l|}{AVE~\cite{tian2018audio}}                              & 49.9  & 37.3 & 37.0 & 41.4 & 43.6  & 43.6 & 32.4 & 32.6 & 36.2 & 37.4  \\
        \multicolumn{2}{l|}{AVSDN~\cite{lin2019dual}}                            & 47.8  & 52.0 & 37.1 & 45.7 & 50.8  & 34.1 & 46.3 & 26.5 & 35.6 & 37.7  \\
        \multicolumn{2}{l|}{HAN~\cite{tian2020unified}}                              & 60.1  & 52.9 & 48.9 & 54.0 & 55.4  & 51.3 & 48.9 & 43.0 & 47.7 & 48.0  \\ \hline
        \multicolumn{2}{l|}{HAN+Co-teaching~\cite{yu2019does}}                  & 59.4  & 56.7 & 52.0 & 56.0 & 56.3  & 50.7 & 53.9 & 46.6 & 50.4 & 48.7  \\
        \multicolumn{2}{l|}{HAN+JoCoR~\cite{wei2020combating}}                        & 61.0  & 58.2 & 53.1 & 57.4 & 57.7  & 52.8 & 54.7 & 46.7 & 51.4 & 50.3  \\ \hline
        \multicolumn{1}{l|}{\multirow{2}{*}{MA~\cite{wu2021exploring}}}     & w/o CL & 59.8  & 57.5 & 52.6 & 56.6 & 56.6  & 52.1 & 54.4 & 45.8 & 50.8 & 49.4  \\
        \multicolumn{1}{l|}{}                        & w CL   & 60.3  & 60.0 & 55.1 & 58.9 & 57.9  & 53.6 & 56.4 & 49.0 & 53.0 & 50.6  \\ \hline
        \multicolumn{1}{l|}{\multirow{2}{*}{JoMoLD~\cite{cheng2022joint}}} & w/o CL & 60.6  & 62.2 & 56.0 & 59.6 & 58.6  & 53.1 & 58.9 & 49.4 & 53.8 & 51.4  \\
        \multicolumn{1}{l|}{}                        & w CL   & 61.3  & 63.8 & 57.2 & 60.8 & 59.9  & 53.9 & 59.9 & 49.6 & 54.5 & 52.5  \\ \hline
        \multicolumn{2}{l|}{MM-Pyramid~\cite{yu2022mm}}                       & 61.1  & 60.3 & 55.8 & 59.7 & 59.1  & 53.8 & 56.7 & 49.4 & 54.1 & 51.2  \\ 
        \multicolumn{2}{l|}{CVCM~\cite{lin2021exploring}}                             & 60.8  & 63.5 & 57.0 & 60.5 & 59.5  & 53.8 & 58.9 & 49.5 & 54.0 & 52.1  \\ \hline
        \multicolumn{2}{l|}{Ours}                             & \textbf{63.5}  & \textbf{65.0} & \textbf{57.8} & \textbf{62.1} & \textbf{62.0}  & \textbf{55.6} & \textbf{61.3} & \textbf{50.5} & \textbf{55.8} & \textbf{54.4}  \\
        \bottomrule[1.2pt]
        \end{tabular}
    }
    \label{tab:sota}
\end{table*}

\subsection{Experimental Settings}

\noindent \textbf{Dataset.} 
Following the standard protocol, we mainly evaluate the performance of our proposed method for WS-AVVP task on LLP
dataset~\cite{tian2020unified}. LLP contains 11849 YouTube video clips spanning over 25 semantic categories, where each video 
clip is 10-second long. 
To perform training, validation and 
testing, we take the common operation to split the dataset into a training set with 10000 videos, a validation 
set with 649 videos and a testing set with 1200 videos. 

To comprehensively investigate the effectiveness of our proposed DGM strategy, we also 
conduct experiments on CREMA-D~\cite{cao2014crema} and AVE~\cite{tian2018audio} datasets: 
CREMA-D is an audio-visual benchmark for speech emotion
recognition. In this dataset, 7442 video clips from 91 actors are collected. This dataset is divided into 
the training and validation subsets, which contain 6698 and 744 videos separately.
AVE is an audio-visual benchmark for audio-visual event localization learning. In this benchmark, there
are 4143 10-second videos from 28 event categories. In our experiments, the split of this dataset follows~\cite{tian2018audio}.

\noindent \textbf{Evaluation Metrics.}
To quantitatively investigate the effectiveness of our proposed method, we evaluate the performance
of our model on all kinds of different events (audio, visual and audio-visual) under both segment-level and event-level
metrics. The segment-level metric is utilized to evaluate the snippet-level prediction performance. In addition, to measure
the event-level F-score results, we concatenate the positive consecutive snippets with the same event category
to produce the event-level prediction, where 0.5 is selected as the mIoU threshold to determine the positive snippets.
Afterwards, the event-level F-score for each event prediction is calculated.
To comprehensively assess the overall performance of our proposed model, the aggregated results are also measured:
Type@AV denotes the average value of audio, visual and audio-visual event localization results. Instead of just averaging
audio, visual and audio-visual metrics, Event@AV measures the results considering all audio and visual events.
For more brevity, in all experimental tables, `A', `V' and `AV' represent the audio, visual and audio-visual events 
respectively. `Type' and `Event' denotes the Type@AV and Event@AV metrics.
On AVE and CREMA-D, we utilize the same evaluation metrics as~\cite{zhou2021positive} and~\cite{peng2022balanced} respectively.

\noindent \textbf{Implementation Details.}
As the standard protocol, given an arbitrary 10-second long video, we first uniformly divide it into a group of 10 non-overlapping snippets, where each snippet contains 8 frames. To extract the visual input, the pre-trained 
ResNet152~\cite{he2016deep} and R(2+1)D~\cite{tran2018closer} networks are utilized to produce the snippet-level appearance and motion feature respectively, which are concatenated at the channel dimension to form the input visual feature. 
In terms of the audio input, a pre-trained VGGish~\cite{hershey2017cnn} model is adopted to extract the audio feature.
In all experiments, we utilize JoMoLD~\cite{cheng2022joint} as the baseline and the training batch size is set to 128. Each model is trained for 25 epochs by using the 
Adam optimizer. $D$ and $\gamma$ are set to $512$ and $0.1$ respectively. 
The initial learning rate for our model is $5e-4$ and drops by a factor of 0.25 for every 6 epochs. 
In the experimens on AVE and CREMA-D, our reproduced PSP~\cite{zhou2021positive} and OGM~\cite{peng2022balanced} methods are utilized as the baselines separately.

\begin{table}
    \centering
    \caption{Effectiveness of DGM on CREMA-D dataset.}
    \resizebox{40mm}{!}{
    \begin{tabular}{l|c}
        \toprule[1.2pt]
        Methods      & Accuracy     \\       \hline
        Baseline~\cite{peng2022balanced}     & 61.1  \\
        DGM   & 61.8 (+0.7)  \\
        \bottomrule[1.2pt]
    \end{tabular}
    }
    \label{tab:discre_CRENAD}
\end{table}

\begin{table}
    \centering
    \caption{Effectiveness of DGM on AVE.}
    \resizebox{40mm}{!}{
    \begin{tabular}{l|c}
        \toprule[1.2pt]
        Methods      & Accuracy \\       \hline
        Baseline~\cite{zhou2021positive}        & 75.57  \\
        DGM   & 76.14 (+0.57) \\
        \bottomrule[1.2pt]
    \end{tabular}
    }
    \label{tab:discre_AVE}
\end{table}

\subsection{Comparison with State-of-the-art Methods}
To comprehensively validate the effectiveness of our model, we compare it with 
different state-of-the-art methods, including the weakly-supervised sound event detection model
TALNet~\cite{wang2019comparison}, weakly-supervised video action detection models CMCS~\cite{liu2019completeness} and STPN~\cite{nguyen2019self}, modified audio-visual 
event localization algorithms AVE~\cite{tian2018audio} and AVSDN~\cite{lin2019dual}, as well as the recent weakly-supervised audio-visual video 
parsing methods HAN~\cite{tian2020unified}, MA~\cite{wu2021exploring}, CVCM~\cite{lin2021exploring}, MM-Pyramid~\cite{yu2022mm} and JoMoLD~\cite{cheng2022joint}. For fair comparison, all
WS-AVVP models are trained by using the LLP training set with the same 
videos and features.

Detailed experimental results of our proposed model and other state-of-the-art methods on the LLP testing subset are 
reported in~\tabref{sota}. Intuitively, our proposed model performs favorably against all other compared 
methods and achieves the best performance on all audio-visual video parsing sub-tasks under both the segment-level and 
event-level evaluation metrics.
Concretely, compared with the most recent JoMoLD model, our proposed model achieves 1.48 average 
performance gains for Audio, Visual, Audio-visual, Type@AV and Event@AV tasks under the segment-level evaluation 
metric. Meanwhile, the average performance improvement for different sub-tasks under the event-level metric is 1.44. 
The above results significantly demonstrate the effectiveness of our proposed method.

Additionally, to comprehensively investigate the effectiveness of our proposed method, 
more experiments are conducted on different audio-visual benchmarks including
CREMA-D and AVE. The corresponding experimental results are summarized in~\tabref{discre_CRENAD} and 
\tabref{discre_AVE}. Consistent performance improvement verifies the effectiveness and good generalization
of our proposed method.

\begin{table}
    \centering
    \caption{Ablation study on the DGM mechanism and MSDU.
    Both segment-level and event-level parsing results are reported. The best results are highlighted in bold.}
    \resizebox{80mm}{!}{%
    \begin{tabular}{c|l|cc}
        \toprule[1.2pt]
        \multicolumn{1}{l|}{Event type} & Methods    & Segment-level & Event-level \\ \hline
        \multirow{3}{*}{A}              & JoMoLD     & 61.3          & 53.9        \\
                                        & JoMoLD+DGM & 61.9          & 54.4        \\
                                        & JoMoLD+DGM+MSDU   & \textbf{63.5} & \textbf{55.6}        \\ \hline
        \multirow{3}{*}{V}              & JoMoLD     & 63.8          & 59.9        \\
                                        & JoMoLD+DGM & 64.4          & 60.9        \\
                                        & JoMoLD+DGM+MSDU   & \textbf{65.0} & \textbf{61.3}        \\ \hline
        \multirow{3}{*}{AV}             & JoMoLD     & 57.2          & 49.6        \\
                                        & JoMoLD+DGM & 56.8          & 49.3        \\
                                        & JoMoLD+DGM+MSDU   & \textbf{57.8} & \textbf{50.5}        \\ \hline
        \multirow{3}{*}{Type}           & JoMoLD     & 60.8          & 54.5        \\
                                        & JoMoLD+DGM & 61.0          & 54.9        \\
                                        & JoMoLD+DGM+MSDU   & \textbf{62.1} & \textbf{55.8}        \\ \hline
        \multirow{3}{*}{Event}          & JoMoLD     & 59.9          & 52.5        \\
                                        & JoMoLD+DGM & 60.6          & 53.4        \\
                                        & JoMoLD+DGM+MSDU   & \textbf{62.0} & \textbf{54.4}        \\ 
        \bottomrule[1.2pt]
    \end{tabular}
    }
    \label{tab:abl_module}
\end{table}

\begin{table}
    \centering
    \caption{Ablation study on different optimization discrepancy measure functions.}
    \resizebox{70mm}{!}{%
    \begin{tabular}{c|l|cc}
        \toprule[1.2pt]
        \multicolumn{1}{l|}{Event type} & Methods      & Segment-level & Event-level \\       \hline
        \multirow{3}{*}{A}              & score        & 63.1          & \textbf{56.0}        \\
                                        & discrepancy  & 62.4          & 55.1        \\
                                        & fusion       & \textbf{63.5} & 55.6        \\ \hline
        \multirow{3}{*}{V}              & score        & 64.4          & 60.9        \\
                                        & discrepancy  & 64.4          & 60.6        \\
                                        & fusion       & \textbf{65.0} & \textbf{61.3}        \\ \hline
        \multirow{3}{*}{AV}             & score        & 57.5          & 49.9        \\
                                        & discrepancy  & 57.0          & 49.6        \\
                                        & fusion       & \textbf{57.8} & \textbf{50.5}        \\ \hline
        \multirow{3}{*}{Type}           & score        & 61.7          & 55.6        \\
                                        & discrepancy  & 61.2          & 55.1        \\
                                        & fusion       & \textbf{62.1} & \textbf{55.8}        \\ \hline
        \multirow{3}{*}{Event}          & score        & 61.7          & \textbf{54.7}        \\
                                        & discrepancy  & 60.9          & 53.8        \\
                                        & fusion       & \textbf{62.0} & 54.4        \\ 
        \bottomrule[1.2pt]
    \end{tabular}
    }
    \label{tab:abl_mea}
\end{table}

\subsection{Ablation Studies}
\label{sec:ablation}

\noindent \textbf{Effectiveness of each component.}
In this part, to further investigate the effectiveness of each component in our method, we conduct comprehensive
ablation studies on LLP dataset. The corresponding experimental results are reported in~\tabref{abl_module}.
From the results, we can observe that our proposed DGM can
consistently improve the audio-visual video parsing performance of different network structures (\emph{i.e.}, JoMoLD and 
our proposed MSDU structure). Concretely, when we insert our proposed DGM mechanism into JoMoLD, the average performance improvements
for Audio, Visual, Type@AV and Event@AV sub-tasks are 0.525 and 0.7 under the Segment-level 
and Event-level evaluation metrics. 
However, the performance for audio-visual sub-task drops slightly, which can be attributed to the factor that 
the confusing multimodal feature will lead to unreliable imbalance measurement.

Compared with `JoMoLD+DGM', MSDU can bring additional 1.14 and 0.94 average performance improvements for different sub-tasks under the Segment-level and 
Event-level metrics respectively. Obviously, our proposed DGM mechanism can bring more performance gains for `JoMoLD+MSDU'
than JoMoLD. We attribute the experimental phenomenon to the reason that the calculations of different modalities 
are confused in JoMoLD, which is not beneficial for pure assessment of imbalanced feature learning between different 
modalities. However, in our designed MSDU block, the calculation of different modalities
is separated and pure, which can boost the effectiveness of DGM.

To sum up, we can draw the following conclusions from the above experimental results:
(1) Our proposed DGM mechanism can improve the audio-visual video parsing 
performance by balancing the feature learning between different modalities; (2) The MSDU 
structure can effectively separate the calculation of different modalities, thus further promote
the effectiveness of our proposed DGM mechanism. (3) Our proposed DGM 
mechanism can be adapted to different backbone structures, achieving consistent performance improvement 
in different WS-AVVP sub-tasks.

\noindent \textbf{Optimization imbalance between different modalities.} 
To reliably measure the imbalanced feature learning between audio and visual modalities, we investigate three different
methods: (1) For each training video, we first calculate the sum of predicted scores for the correct categories in audio and visual 
modalities separately. Afterwards, the ratio of the summed predicted scores is calculated as the imbalance degree between 
two modalities. (2) We calculate the discrepancies between the average predicted scores for the correct and false semantic 
categories in different modalities firstly. Thereafter, the ratio of the discrepancies in different modalities is utilized to 
measure the imbalanced feature learning between audio and visual modalities. (3) The above two methods are combined together to 
assess the imbalanced optimization between audio and visual feature encoders.
Detailed experimental results are listed in~\tabref{abl_mea}. `score', `discrepancy' and `fusion' denote the three 
different methods mentioned above.
From the experimental results, we can conclude that all three methods can measure the optimization imbalance between
audio and visual feature encoders effectively.
When we combine the two different information utilized in the first and second strategies, our DGM mechanism can achieve the 
most performance gains (\emph{i.e.,} `JoMoLD+DGM+MSDU' in~\tabref{abl_module}). 
We attribute this phenomenon to the reason that `score' strategy only takes the predictions for the correct 
categories into consideration, which neglects the global distribution of the classification predictions. Similarly, `discrepancy'
method only considers the discrepancy between the predictions for correct and false event categories while ignores the original
prediction scores for the correct categories. When we combine the above two strategies, more exhaustive information 
will provide more reliable assessment of the optimization imbalance between different feature encoders, which will better 
balance the multimodal feature learning and improve the audio-visual video paring performance.

\begin{table*}
    \centering
    \caption{Ablation study on the effectiveness of different $\gamma$ in Eq.(7) of our main paper. Bold and underlined represent the optimal and sub-optimal performance.}
    \resizebox{135mm}{!}{%
    \begin{tabular}{cc|ccccc|ccccc|c}
        \toprule[1.2pt]
        \multicolumn{2}{c|}{\multirow{2}{*}{$\gamma$}}        & \multicolumn{5}{c|}{Segment-level} & \multicolumn{5}{c|}{Event-level} & {\multirow{2}{*}{AVG}}  \\ \cline{3-12} 
        \multicolumn{2}{l|}{}                                 & A     & V    & AV   & Type & Event & A    & V    & AV   & Type & Event \\ \hline
        \multicolumn{2}{l|}{0.1}                              & \underline{63.5}  & \underline{65.0} & \textbf{57.8} & \textbf{62.1} & 62.0  & 55.6 & \underline{61.3} & \textbf{50.5} & \textbf{55.8} & 54.4 & \textbf{58.80}  \\
        \multicolumn{2}{l|}{0.2}                              & \textbf{63.6}  & 64.5 & \underline{57.7} & 61.9 & 62.0  & \textbf{56.1} & 60.8 & \underline{50.2} & \underline{55.7} & \textbf{54.6} & \underline{58.71} \\
        \multicolumn{2}{l|}{0.3}                              & 63.2  & \textbf{65.4} & 57.6 & \textbf{62.1} & 61.9  & 55.2 & \textbf{61.5} & 50.0 & 55.6 & 54.1 & 58.66 \\
        \multicolumn{2}{l|}{0.4}                              & \textbf{63.6}  & 64.8 & 57.6 & \underline{62.0} & \underline{62.1}  & \underline{55.7} & 60.7 & 49.8 & 55.4 & 54.2 & 58.59 \\ 
        \multicolumn{2}{l|}{0.5}                              & \textbf{63.6}  & 64.8 & 57.3 & 61.9 & \textbf{62.2}  & \underline{55.7} & 60.9 & 49.5 & 55.3 & \underline{54.5} & 58.57 \\
        \multicolumn{2}{l|}{0.6}                              & 62.9  & 64.9 & 57.6 & 61.8 & 61.7  & 54.9 & \underline{61.3} & 49.9 & 55.4 & 53.9 & 58.43 \\
        \multicolumn{2}{l|}{0.7}                              & 62.9  & 64.5 & 57.5 & 61.7 & 61.2  & 55.5 & 60.5 & 49.9 & 55.3 & 53.8 & 58.28 \\ 
        \multicolumn{2}{l|}{0.8}                              & 62.6  & \underline{65.0} & 57.3 & 61.6 & 61.1  & 54.5 & 61.1 & 49.7 & 55.1 & 53.2 & 58.12 \\
        \multicolumn{2}{l|}{0.9}                              & 62.7  & 64.5 & 57.0 & 61.4 & 61.6  & 55.1 & 60.9 & 49.4 & 55.2 & 54.1 & 58.19 \\ 
        \bottomrule[1.2pt]
        \end{tabular}
    }
    \label{tab:gamma}
 \end{table*}

 \begin{figure*}
    \subfigure[Video `9LNuWqOQP5Q']{
      \includegraphics[width=0.95\linewidth]{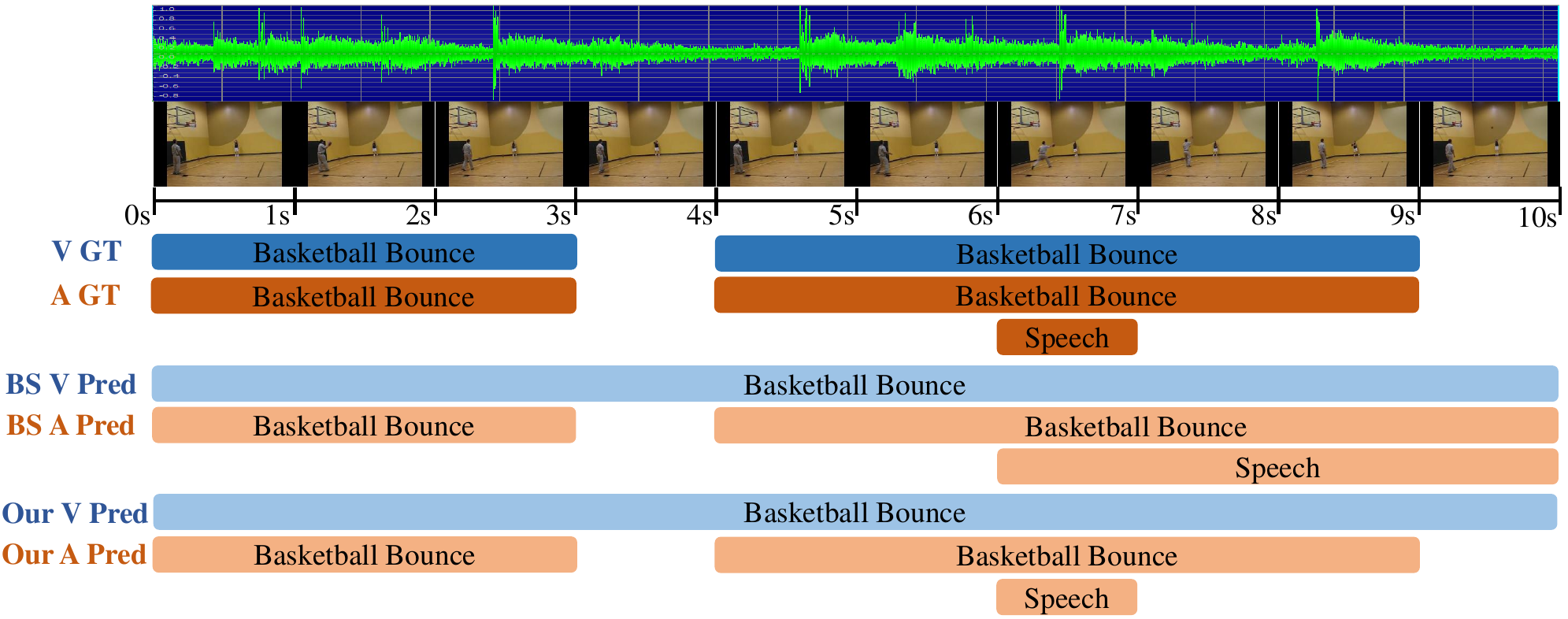}
    }
    \subfigure[Video `K9KdTtdCRWw']{
      \includegraphics[width=0.95\linewidth]{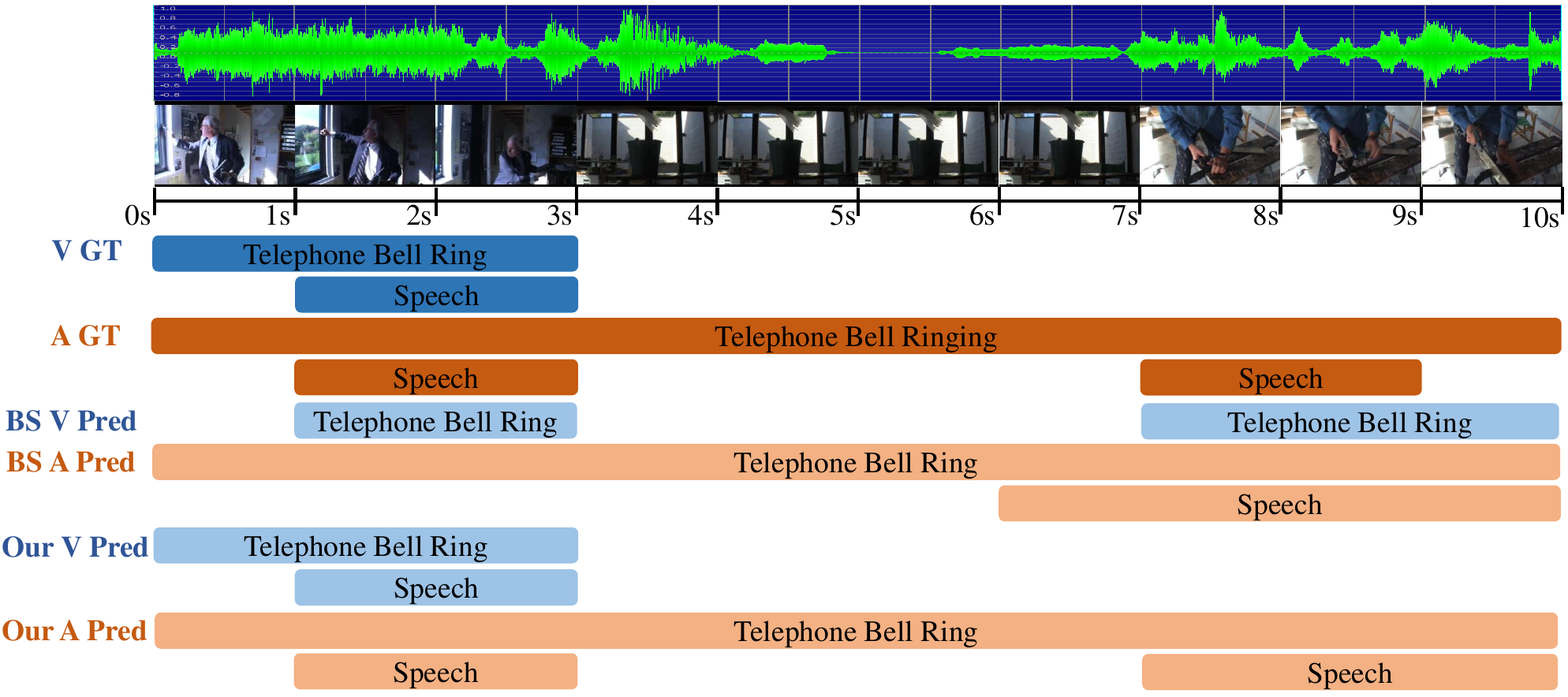}
    }
    \subfigure[Video `Mzp5uTDaKog']{
        \includegraphics[width=0.95\linewidth]{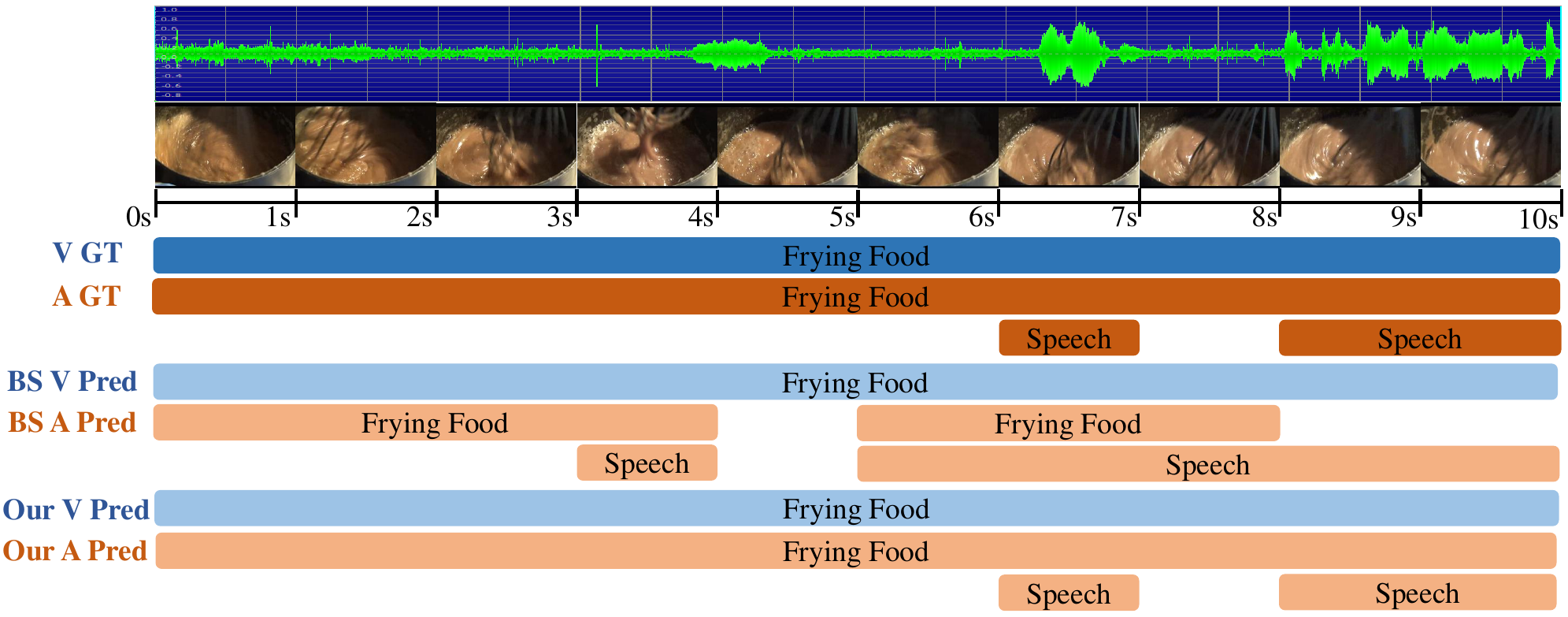}
    }
    \caption{Illustration of groundtruths as well as the predictions of the baseline method and our proposed whole model.
    `V GT' and `A GT' denote visual and audio event groundtruths, `BS' indicates the baseline structure, 
    and `Our' is our proposed method (\emph{i.e.}, JoMoLD equipped with our proposed MSDU and DGM).}
    \label{fig:visualization}
\end{figure*}


\begin{figure}
    \begin{center}
        {\includegraphics[width=0.98\linewidth]{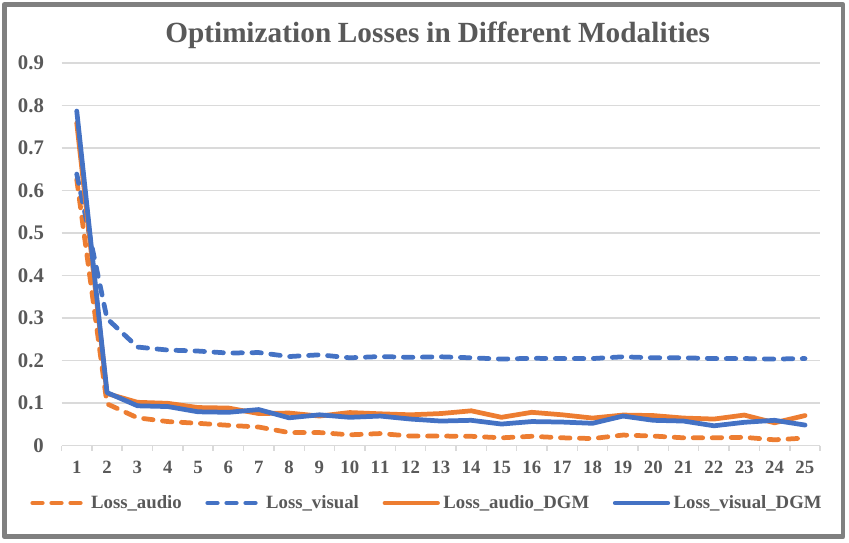}}
    \end{center}
    \caption{Losses in the training processes before and after the gradients in different modalities are modulated by our proposed
    DGM mechanism. x-coordinate and y-coordinate indicate the training epoch and loss value respectively.
    Dashed and solid lines denote the losses before and after the gradients are modulated by our proposed DGM mechanism, and different colors 
    represent the different modalities.}
    \label{fig:loss}
\end{figure}

\begin{table*}
    \centering
    \caption{Ablation study on the generalization of our proposed method in different WS-AVVP models.}
    \resizebox{135mm}{!}{%
    \begin{tabular}{cc|ccccc|ccccc|c}
        \toprule[1.2pt]
        \multicolumn{2}{c|}{\multirow{2}{*}{Method}}        & \multicolumn{5}{c|}{Segment-level} & \multicolumn{5}{c|}{Event-level} & {\multirow{2}{*}{AVG}}  \\ \cline{3-12} 
        \multicolumn{2}{l|}{}                               & A     & V    & AV   & Type & Event & A    & V    & AV   & Type & Event \\ \hline
        \multicolumn{2}{l|}{HAN}                            & 60.1  & 52.9  &  48.9 &  54.0 &  55.4 &  51.3 & 48.9  & 43.0  &  47.7 & 48.0  &  51.02  \\
        \multicolumn{2}{l|}{HAN + Ours}                     & 61.0  & 55.2  &  51.5 &  55.9 &  56.2 &  53.4 & 51.0  & 45.1  &  49.8 & 50.0  &  52.91  \\
        \hline
        \multicolumn{2}{l|}{JoMoLD}                         & 61.3  & 63.8  &  57.2 &  60.8 &  59.9 &  53.9 & 59.9  & 49.6  &  54.5 & 52.5  &  57.34  \\
        \multicolumn{2}{l|}{JoMoLD + Ours}                  & 63.5  & 65.0  &  57.8 &  62.1 &  62.0 &  55.6 & 61.3  & 50.5  &  55.8 & 54.4  &  58.80  \\ 
        \bottomrule[1.2pt]
        \end{tabular}
    }
    \label{tab:generalization}
 \end{table*}

\noindent \textbf{Effectiveness of $\gamma$ in Eq.(7) of our main paper.}
To analyze the effect of $\gamma$ in Eq.(7) on our proposed DGM mechanism, we conduct comprehensive ablations and the corresponding
results are reported in~\tabref{gamma}. From the experimental results, we can conclude that when $\gamma$ is equal to 0.1, our model 
achieves the best audio-visual video parsing performance.
Additionally, when $\gamma$ increases from 0.2 to 0.9, our model performance decreases slightly.
However, compared with `JoMoLD' model in Table (4) of our main paper, our proposed DGM mechanism can always improve the performance no matter what $\gamma$ is, 
which verifies that our proposed DGM can robustly balance the optimization processes of different modalities 
and further improve the performance of our proposed WS-AVVP model.

\noindent \textbf{Balanced optimization between different modalities.}
To intuitively verify that our proposed DGM mechanism can balance the optimization between audio and visual modalities, 
we analyze the training losses of different modalities before and after the gradients are modulated by our proposed 
DGM mechanism. As illustrated in~\figref{loss}, x-coordinate and y-coordinate denote the training epoch and the corresponding loss respectively.
Intuitively, we can observe that there is a large gap between the losses of audio and visual modalities before the gradients
of our designed WS-AVVP model are modulated. Meanwhile, after audio and visual sub-networks are modulated by our proposed
DGM, the traning losses of different modalities are balanced, which strongly proves that our proposed DGM mechanism can make the 
suboptimal modality achieve more optimization attention and further balance the feature learning processes of different modalities.

\noindent \textbf{Generalization of our proposed method.}
To verify the generalization ability of our proposed DGM strategy, we embed it into different WS-AVVP 
models including HAN~\cite{tian2020unified} and JoMoLD~\cite{cheng2022joint}. The corresponding experimental results are summarized in~\tabref{generalization}. 
Obviously, our proposed method can consistently improve the audio-visual video parsing performance of 
differnt baseline models, which proves that our method can cooperate with variant model structures for 
better AVVP performance without being limited by the specific model structure.

\noindent \textbf{Qualitative Results.}
To investigate the effectiveness of our proposed method more intuitively, we compare the qualitative audio-visual parsing
predictions produced by the baseline and our proposed complete model (\emph{i.e.}, JoMoLD+MSDU+DGM) in~\figref{visualization}. 
From the results, we can conclude that our 
complete method can localize more accurate audio, visual, audio-visual event instances than baseline structure, which 
verifies that our proposed DGM can effectively improve the WS-AVVP performance.
In addition, we also observe that compared with the baseline model, our proposed DGM mechanism can help 
detect more precise event instances in the relatively weak modality. 
For example, as shown in subfigure (c) of~\figref{visualization}, it is hard to judge whether `Frying Food'
is happening by just listening to the audio, but it is obvious in visual modality. 
Consequently, the bseline method can only detect the accurate event instances in visual modality but not
in audio. However, our proposed method can not only precisely localize the events
in visual modality but also in audio.
Consequently, we can conclude that our DGM can improve the video parsing performance in relatively weak modalities, 
which further proves that our proposed DGM can balance the optimization processes between different modalities and 
make the weak modalities achieve more optimization attention.
\section{Conclusion}
\label{sec:conclusion}
In this paper, we first analyze the imbalanced feature learning between different modalities 
in WS-AVVP task. To mitigate the above issue, the dynamic 
gradient modulation strategy is designed to modulate the gradients of the feature encoders 
for different modalities,
so as to make the model pay more optimization attention to the suboptimal branch.
Meanwhile, to address the negative effect caused by the confusing multimodal feature 
on our proposed DGM, 
we design a modality-separated decision unit for more precise measurement of the imbalanced feature 
learning between audio and visual modalities. 
Comprehensive experiments verify the effectiveness of our proposed method.

\bibliographystyle{IEEEtran}
\small
\bibliography{IEEEabrv,sample-base}

\begin{thebibliography}{10}
\providecommand{\url}[1]{#1}
\csname url@samestyle\endcsname
\providecommand{\newblock}{\relax}
\providecommand{\bibinfo}[2]{#2}
\providecommand{\BIBentrySTDinterwordspacing}{\spaceskip=0pt\relax}
\providecommand{\BIBentryALTinterwordstretchfactor}{4}
\providecommand{\BIBentryALTinterwordspacing}{\spaceskip=\fontdimen2\font plus
\BIBentryALTinterwordstretchfactor\fontdimen3\font minus
  \fontdimen4\font\relax}
\providecommand{\BIBforeignlanguage}[2]{{%
\expandafter\ifx\csname l@#1\endcsname\relax
\typeout{** WARNING: IEEEtran.bst: No hyphenation pattern has been}%
\typeout{** loaded for the language `#1'. Using the pattern for}%
\typeout{** the default language instead.}%
\else
\language=\csname l@#1\endcsname
\fi
#2}}
\providecommand{\BIBdecl}{\relax}
\BIBdecl

\bibitem{planamente2022domain}
M.~Planamente, C.~Plizzari, E.~Alberti, and B.~Caputo, ``Domain generalization
  through audio-visual relative norm alignment in first person action
  recognition,'' in \emph{Proceedings of the IEEE/CVF Winter Conference on
  Applications of Computer Vision}, 2022, pp. 1807--1818.

\bibitem{gao2020listen}
R.~Gao, T.-H. Oh, K.~Grauman, and L.~Torresani, ``Listen to look: Action
  recognition by previewing audio,'' in \emph{Proceedings of the IEEE/CVF
  Conference on Computer Vision and Pattern Recognition}, 2020, pp.
  10\,457--10\,467.

\bibitem{kazakos2019epic}
E.~Kazakos, A.~Nagrani, A.~Zisserman, and D.~Damen, ``Epic-fusion: Audio-visual
  temporal binding for egocentric action recognition,'' in \emph{Proceedings of
  the IEEE/CVF International Conference on Computer Vision}, 2019, pp.
  5492--5501.

\bibitem{tzinis2022audioscopev2}
E.~Tzinis, S.~Wisdom, T.~Remez, and J.~R. Hershey, ``Audioscopev2: Audio-visual
  attention architectures for calibrated open-domain on-screen sound
  separation,'' in \emph{European Conference on Computer Vision}.\hskip 1em
  plus 0.5em minus 0.4em\relax Springer, 2022, pp. 368--385.

\bibitem{gan2020music}
C.~Gan, D.~Huang, H.~Zhao, J.~B. Tenenbaum, and A.~Torralba, ``Music gesture
  for visual sound separation,'' in \emph{Proceedings of the IEEE/CVF
  Conference on Computer Vision and Pattern Recognition}, 2020, pp.
  10\,478--10\,487.

\bibitem{tzinis2020into}
E.~Tzinis, S.~Wisdom, A.~Jansen, S.~Hershey, T.~Remez, D.~Ellis, and J.~R.
  Hershey, ``Into the wild with audioscope: Unsupervised audio-visual
  separation of on-screen sounds,'' in \emph{International Conference on
  Learning Representations}, 2020.

\bibitem{tian2021cyclic}
Y.~Tian, D.~Hu, and C.~Xu, ``Cyclic co-learning of sounding object visual
  grounding and sound separation,'' in \emph{Proceedings of the IEEE/CVF
  Conference on Computer Vision and Pattern Recognition}, 2021, pp. 2745--2754.

\bibitem{gao2021visualvoice}
R.~Gao and K.~Grauman, ``Visualvoice: Audio-visual speech separation with
  cross-modal consistency,'' in \emph{2021 IEEE/CVF Conference on Computer
  Vision and Pattern Recognition (CVPR)}.\hskip 1em plus 0.5em minus
  0.4em\relax IEEE, 2021, pp. 15\,490--15\,500.

\bibitem{tian2018audio}
Y.~Tian, J.~Shi, B.~Li, Z.~Duan, and C.~Xu, ``Audio-visual event localization
  in unconstrained videos,'' in \emph{Proceedings of the European Conference on
  Computer Vision (ECCV)}, 2018, pp. 247--263.

\bibitem{wu2019dual}
Y.~Wu, L.~Zhu, Y.~Yan, and Y.~Yang, ``Dual attention matching for audio-visual
  event localization,'' in \emph{Proceedings of the IEEE/CVF international
  conference on computer vision}, 2019, pp. 6292--6300.

\bibitem{xuan2020atten}
H.~Xuan, Z.~Zhang, S.~Chen, J.~Yang, and Y.~Yan, ``Cross-modal attention
  network for temporal inconsistent audio-visual event localization,'' in
  \emph{Proceedings of the AAAI Conference on Artificial Intelligence}, 2020,
  pp. 279--286.

\bibitem{duan2021audio}
B.~Duan, H.~Tang, W.~Wang, Z.~Zong, G.~Yang, and Y.~Yan, ``Audio-visual event
  localization via recursive fusion by joint co-attention,'' in
  \emph{Proceedings of the IEEE/CVF Winter Conference on Applications of
  Computer Vision}, 2021, pp. 4013--4022.

\bibitem{xue2021audio}
C.~Xue, X.~Zhong, M.~Cai, H.~Chen, and W.~Wang, ``Audio-visual event
  localization by learning spatial and semantic co-attention,'' \emph{IEEE
  Transactions on Multimedia}, 2021.

\bibitem{xu2020cross}
H.~Xu, R.~Zeng, Q.~Wu, M.~Tan, and C.~Gan, ``Cross-modal relation-aware
  networks for audio-visual event localization,'' in \emph{Proceedings of the
  28th ACM International Conference on Multimedia}, 2020, pp. 3893--3901.

\bibitem{tian2020unified}
Y.~Tian, D.~Li, and C.~Xu, ``Unified multisensory perception: Weakly-supervised
  audio-visual video parsing,'' in \emph{European Conference on Computer
  Vision}.\hskip 1em plus 0.5em minus 0.4em\relax Springer, 2020, pp. 436--454.

\bibitem{lin2021exploring}
Y.-B. Lin, H.-Y. Tseng, H.-Y. Lee, Y.-Y. Lin, and M.-H. Yang, ``Exploring
  cross-video and cross-modality signals for weakly-supervised audio-visual
  video parsing,'' \emph{Advances in Neural Information Processing Systems},
  vol.~34, pp. 11\,449--11\,461, 2021.

\bibitem{cheng2022joint}
H.~Cheng, Z.~Liu, H.~Zhou, C.~Qian, W.~Wu, and L.~Wang, ``Joint-modal label
  denoising for weakly-supervised audio-visual video parsing,'' \emph{European
  Conference on Computer Vision}, 2022.

\bibitem{parida2020coordinated}
K.~Parida, N.~Matiyali, T.~Guha, and G.~Sharma, ``Coordinated joint multimodal
  embeddings for generalized audio-visual zero-shot classification and
  retrieval of videos,'' in \emph{Proceedings of the IEEE/CVF Winter Conference
  on Applications of Computer Vision}, 2020, pp. 3251--3260.

\bibitem{ismail2020improving}
A.~A. Ismail, M.~Hasan, and F.~Ishtiaq, ``Improving multimodal accuracy through
  modality pre-training and attention,'' \emph{arXiv preprint
  arXiv:2011.06102}, 2020.

\bibitem{sun2021learning}
Y.~Sun, S.~Mai, and H.~Hu, ``Learning to balance the learning rates between
  various modalities via adaptive tracking factor,'' \emph{IEEE Signal
  Processing Letters}, vol.~28, pp. 1650--1654, 2021.

\bibitem{wang2020makes}
W.~Wang, D.~Tran, and M.~Feiszli, ``What makes training multi-modal
  classification networks hard?'' in \emph{Proceedings of the IEEE/CVF
  Conference on Computer Vision and Pattern Recognition}, 2020, pp.
  12\,695--12\,705.

\bibitem{peng2022balanced}
X.~Peng, Y.~Wei, A.~Deng, D.~Wang, and D.~Hu, ``Balanced multimodal learning
  via on-the-fly gradient modulation,'' in \emph{Proceedings of the IEEE/CVF
  Conference on Computer Vision and Pattern Recognition}, 2022, pp. 8238--8247.

\bibitem{aytar2016soundnet}
Y.~Aytar, C.~Vondrick, and A.~Torralba, ``Soundnet: Learning sound
  representations from unlabeled video,'' \emph{Advances in neural information
  processing systems}, vol.~29, 2016.

\bibitem{hu2019deep}
D.~Hu, F.~Nie, and X.~Li, ``Deep multimodal clustering for unsupervised
  audiovisual learning,'' in \emph{Proceedings of the IEEE/CVF Conference on
  Computer Vision and Pattern Recognition}, 2019, pp. 9248--9257.

\bibitem{korbar2018cooperative}
B.~Korbar, D.~Tran, and L.~Torresani, ``Cooperative learning of audio and video
  models from self-supervised synchronization,'' \emph{Advances in Neural
  Information Processing Systems}, vol.~31, 2018.

\bibitem{arandjelovic2017look}
R.~Arandjelovic and A.~Zisserman, ``Look, listen and learn,'' in
  \emph{Proceedings of the IEEE International Conference on Computer Vision},
  2017, pp. 609--617.

\bibitem{cheng2020look}
Y.~Cheng, R.~Wang, Z.~Pan, R.~Feng, and Y.~Zhang, ``Look, listen, and attend:
  Co-attention network for self-supervised audio-visual representation
  learning,'' in \emph{Proceedings of the 28th ACM International Conference on
  Multimedia}, 2020, pp. 3884--3892.

\bibitem{ma2020active}
S.~Ma, Z.~Zeng, D.~McDuff, and Y.~Song, ``Active contrastive learning of
  audio-visual video representations,'' in \emph{International Conference on
  Learning Representations}, 2020.

\bibitem{morgado2020learning}
P.~Morgado, Y.~Li, and N.~Nvasconcelos, ``Learning representations from
  audio-visual spatial alignment,'' \emph{Advances in Neural Information
  Processing Systems}, vol.~33, pp. 4733--4744, 2020.

\bibitem{rachavarapu2021localize}
K.~K. Rachavarapu, V.~Sundaresha, A.~Rajagopalan \emph{et~al.}, ``Localize to
  binauralize: Audio spatialization from visual sound source localization,'' in
  \emph{Proceedings of the IEEE/CVF International Conference on Computer
  Vision}, 2021, pp. 1930--1939.

\bibitem{xuan2022proposal}
H.~Xuan, Z.~Wu, J.~Yang, Y.~Yan, and X.~Alameda-Pineda, ``A proposal-based
  paradigm for self-supervised sound source localization in videos,'' in
  \emph{Proceedings of the IEEE/CVF Conference on Computer Vision and Pattern
  Recognition}, 2022, pp. 1029--1038.

\bibitem{owens2018audio}
A.~Owens and A.~A. Efros, ``Audio-visual scene analysis with self-supervised
  multisensory features,'' in \emph{Proceedings of the European Conference on
  Computer Vision (ECCV)}, 2018, pp. 631--648.

\bibitem{senocak2018learning}
A.~Senocak, T.-H. Oh, J.~Kim, M.-H. Yang, and I.~S. Kweon, ``Learning to
  localize sound source in visual scenes,'' in \emph{Proceedings of the IEEE
  Conference on Computer Vision and Pattern Recognition}, 2018, pp. 4358--4366.

\bibitem{hori2017attention}
C.~Hori, T.~Hori, T.-Y. Lee, Z.~Zhang, B.~Harsham, J.~R. Hershey, T.~K. Marks,
  and K.~Sumi, ``Attention-based multimodal fusion for video description,'' in
  \emph{Proceedings of the IEEE international conference on computer vision},
  2017, pp. 4193--4202.

\bibitem{rahman2019watch}
T.~Rahman, B.~Xu, and L.~Sigal, ``Watch, listen and tell: Multi-modal weakly
  supervised dense event captioning,'' in \emph{Proceedings of the IEEE/CVF
  international conference on computer vision}, 2019, pp. 8908--8917.

\bibitem{tian2018attempt}
Y.~Tian, C.~Guan, J.~Goodman, M.~Moore, and C.~Xu, ``An attempt towards
  interpretable audio-visual video captioning,'' \emph{arXiv preprint
  arXiv:1812.02872}, 2018.

\bibitem{gan2019self}
C.~Gan, H.~Zhao, P.~Chen, D.~Cox, and A.~Torralba, ``Self-supervised moving
  vehicle tracking with stereo sound,'' in \emph{Proceedings of the IEEE/CVF
  International Conference on Computer Vision}, 2019, pp. 7053--7062.

\bibitem{zhao2019sound}
H.~Zhao, C.~Gan, W.-C. Ma, and A.~Torralba, ``The sound of motions,'' in
  \emph{Proceedings of the IEEE/CVF International Conference on Computer
  Vision}, 2019, pp. 1735--1744.

\bibitem{zhao2018sound}
H.~Zhao, C.~Gan, A.~Rouditchenko, C.~Vondrick, J.~McDermott, and A.~Torralba,
  ``The sound of pixels,'' in \emph{Proceedings of the European conference on
  computer vision (ECCV)}, 2018, pp. 570--586.

\bibitem{wu2021exploring}
Y.~Wu and Y.~Yang, ``Exploring heterogeneous clues for weakly-supervised
  audio-visual video parsing,'' in \emph{Proceedings of the IEEE/CVF Conference
  on Computer Vision and Pattern Recognition}, 2021, pp. 1326--1335.

\bibitem{du2021improving}
C.~Du, T.~Li, Y.~Liu, Z.~Wen, T.~Hua, Y.~Wang, and H.~Zhao, ``Improving
  multi-modal learning with uni-modal teachers,'' \emph{arXiv preprint
  arXiv:2106.11059}, 2021.

\bibitem{mandt2017stochastic}
S.~Mandt, M.~D. Hoffman, and D.~M. Blei, ``Stochastic gradient descent as
  approximate bayesian inference,'' \emph{arXiv preprint arXiv:1704.04289},
  2017.

\bibitem{jastrzkebski2017three}
S.~Jastrzebski, Z.~Kenton, D.~Arpit, N.~Ballas, A.~Fischer, Y.~Bengio, and
  A.~Storkey, ``Three factors influencing minima in sgd,'' \emph{arXiv preprint
  arXiv:1711.04623}, 2017.

\bibitem{yu2019does}
X.~Yu, B.~Han, J.~Yao, G.~Niu, I.~Tsang, and M.~Sugiyama, ``How does
  disagreement help generalization against label corruption?'' in
  \emph{International Conference on Machine Learning}.\hskip 1em plus 0.5em
  minus 0.4em\relax PMLR, 2019, pp. 7164--7173.

\bibitem{wei2020combating}
H.~Wei, L.~Feng, X.~Chen, and B.~An, ``Combating noisy labels by agreement: A
  joint training method with co-regularization,'' in \emph{Proceedings of the
  IEEE/CVF Conference on Computer Vision and Pattern Recognition}, 2020, pp.
  13\,726--13\,735.

\bibitem{wang2019comparison}
Y.~Wang, J.~Li, and F.~Metze, ``A comparison of five multiple instance learning
  pooling functions for sound event detection with weak labeling,'' in
  \emph{ICASSP 2019-2019 IEEE International Conference on Acoustics, Speech and
  Signal Processing (ICASSP)}.\hskip 1em plus 0.5em minus 0.4em\relax IEEE,
  2019, pp. 31--35.

\bibitem{nguyen2019self}
D.~T. Nguyen, C.~K. Mummadi, T.~P.~N. Ngo, T.~H.~P. Nguyen, L.~Beggel, and
  T.~Brox, ``Self: Learning to filter noisy labels with self-ensembling,''
  \emph{arXiv preprint arXiv:1910.01842}, 2019.

\bibitem{liu2019completeness}
D.~Liu, T.~Jiang, and Y.~Wang, ``Completeness modeling and context separation
  for weakly supervised temporal action localization,'' in \emph{Proceedings of
  the IEEE/CVF Conference on Computer Vision and Pattern Recognition}, 2019,
  pp. 1298--1307.

\bibitem{lin2019dual}
Y.-B. Lin, Y.-J. Li, and Y.-C.~F. Wang, ``Dual-modality seq2seq network for
  audio-visual event localization,'' in \emph{ICASSP 2019-2019 IEEE
  International Conference on Acoustics, Speech and Signal Processing
  (ICASSP)}.\hskip 1em plus 0.5em minus 0.4em\relax IEEE, 2019, pp. 2002--2006.

\bibitem{yu2022mm}
J.~Yu, Y.~Cheng, R.-W. Zhao, R.~Feng, and Y.~Zhang, ``Mm-pyramid: multimodal
  pyramid attentional network for audio-visual event localization and video
  parsing,'' in \emph{Proceedings of the 30th ACM International Conference on
  Multimedia}, 2022, pp. 6241--6249.

\bibitem{cao2014crema}
H.~Cao, D.~G. Cooper, M.~K. Keutmann, R.~C. Gur, A.~Nenkova, and R.~Verma,
  ``Crema-d: Crowd-sourced emotional multimodal actors dataset,'' \emph{IEEE
  transactions on affective computing}, vol.~5, no.~4, pp. 377--390, 2014.

\bibitem{zhou2021positive}
J.~Zhou, L.~Zheng, Y.~Zhong, S.~Hao, and M.~Wang, ``Positive sample propagation
  along the audio-visual event line,'' in \emph{Proceedings of the IEEE/CVF
  Conference on Computer Vision and Pattern Recognition}, 2021, pp. 8436--8444.

\bibitem{he2016deep}
K.~He, X.~Zhang, S.~Ren, and J.~Sun, ``Deep residual learning for image
  recognition,'' in \emph{Proceedings of the IEEE conference on computer vision
  and pattern recognition}, 2016, pp. 770--778.

\bibitem{tran2018closer}
D.~Tran, H.~Wang, L.~Torresani, J.~Ray, Y.~LeCun, and M.~Paluri, ``A closer
  look at spatiotemporal convolutions for action recognition,'' in \emph{IEEE
  Conference on Computer Vision and Pattern Recognition}, 2018.

\bibitem{hershey2017cnn}
S.~Hershey, S.~Chaudhuri, D.~P. Ellis, J.~F. Gemmeke, A.~Jansen, R.~C. Moore,
  M.~Plakal, D.~Platt, R.~A. Saurous, B.~Seybold \emph{et~al.}, ``Cnn
  architectures for large-scale audio classification,'' in \emph{2017 ieee
  international conference on acoustics, speech and signal processing
  (icassp)}.\hskip 1em plus 0.5em minus 0.4em\relax IEEE, 2017, pp. 131--135.

\end{thebibliography}


\end{document}